%% file: main.tex
\documentclass[10pt,twocolumn,letterpaper]{article}

\usepackage{iccv}
\usepackage{times}
\usepackage{epsfig}
\usepackage{graphicx}
\usepackage{amsmath}
\usepackage{amssymb}

\usepackage[T1]{fontenc}    

\usepackage{url}            
\usepackage{booktabs}       
\usepackage{amsfonts}       
\usepackage{nicefrac}       
\usepackage{microtype}      
\usepackage[dvipsnames]{xcolor}
\usepackage{adjustbox}
\usepackage{bm}
\usepackage{amsmath}
\usepackage{subfigure}
\usepackage{wrapfig}
\usepackage{multirow}
\usepackage{bbm}

\usepackage{amssymb}
\usepackage{pifont}
\newcommand{\cmark}{\ding{51}}%
\newcommand{\xmark}{\ding{55}}%
\usepackage[accsupp]{axessibility} 

\usepackage{url}
\def\vs{\emph{vs}~}

\def\methodname{{CoinSeg}~}

\usepackage[pagebackref=true,breaklinks=true,letterpaper=true,colorlinks,bookmarks=false,colorlinks,linkcolor={Mahogany},citecolor={YellowGreen},urlcolor={CadetBlue}]{hyperref}

\iccvfinalcopy 

\def\iccvPaperID{5399} 

\ificcvfinal\fi

\begin{document}


\title{CoinSeg: Contrast Inter- and Intra- Class Representations \\for Incremental Segmentation}
\author{%
  Zekang Zhang$^{1}$\thanks{Work done during an intern at WEI Lab of Beijing Jiaotong University.},\quad Guangyu Gao$^{1}$\thanks{Corresponding author. \textit{guangyugao@bit.edu.cn}.}, \quad Jianbo Jiao$^{2}$, \quad Chi Harold Liu$^{1}$, \quad Yunchao Wei$^{3, 4}$\\
  \\
  $^1$~School of Computer Science, Beijing Institute of Technology\\
  $^2$~School of Computer Science, University of Birmingham\\
  $^3$~WEI Lab, Institute of Information Science, Beijing Jiaotong University \\
  $^{4}$~Beijing Key Laboratory of Advanced Information Science and Network \\
  \texttt{zkzhang1998@outlook.com} \\
}

\maketitle
\ificcvfinal\thispagestyle{empty}\fi


\input{sections/0_abstract.tex}


\input{sections/1_introduction-gao}

\input{sections/2_related_work.tex}

\input{sections/3_method.tex}

\input{sections/4_experiment.tex}
\newpage
\input{sections/5_conclusion.tex}

\newpage

{\small
\bibliographystyle{ieee_fullname}
\bibliography{reference}
}

\end{document}

%% file: sections/0_abstract.tex
\begin{abstract}
Class incremental semantic segmentation aims to strike a balance between the model’s stability and plasticity by maintaining old knowledge while adapting to new concepts. 
However, most state-of-the-art methods use the freeze strategy for stability, which compromises the model's plasticity.
In contrast, releasing parameter training for plasticity could lead to the best performance for all categories, but this requires discriminative feature representation.
Therefore, we prioritize the model's plasticity and propose the \textbf{Co}ntrast \textbf{in}ter- and \textbf{in}tra-class representations for Incremental \textbf{Seg}mentation~(\textbf{CoinSeg}), which pursues discriminative representations for flexible parameter tuning.
Inspired by the Gaussian mixture model that samples from a mixture of Gaussian distributions, CoinSeg emphasizes intra-class diversity with multiple contrastive representation centroids.
Specifically, we use mask proposals to identify regions with strong objectness that are likely to be diverse instances/centroids of a category.
These mask proposals are then used for contrastive representations to reinforce intra-class diversity.
Meanwhile, to avoid bias from intra-class diversity, we also apply category-level pseudo-labels to enhance category-level consistency and inter-category diversity.
Additionally, CoinSeg ensures the model's stability and alleviates forgetting through a specific flexible tuning strategy.
We validate CoinSeg on Pascal VOC 2012 and ADE20K datasets with multiple incremental scenarios and achieve superior results compared to previous state-of-the-art methods, especially in more challenging and realistic long-term scenarios. Code is available at 
\href{https://github.com/zkzhang98/CoinSeg}{https://github.com/zkzhang98/CoinSeg}.
\end{abstract}

%% file: sections/1_introduction-gao.tex
\section{Introduction}
\input{figures/fig1_PMM}
In recent years, deep learning based methods have achieved satisfactory performance on various recognition tasks, with the assumption of fixed or stable data distribution~\cite{lecun2015deep}.
However, real-world data is typically a continuous stream with an unstable distribution, making it difficult for models to retain old knowledge while acquiring new concepts, known as \textit{catastrophic forgetting}~\cite{forget0_cauwenberghs2000incremental,forget_mccloskey1989catastrophic,forget1_polikar2001learn++}.
To tackle this problem, \textit{incremental learning} is proposed to adapt to changing data streams for new concepts, but also avoid forgetting old knowledge, especially for the classification task, \ie, Class-Incremental Learning~(CIL)~\cite{LFL_jung2016less,LwF_li2017learning, zhang2023slca}.

Class Incremental Semantic Segmentation(CISS) aims to assign an image with the pixel-wise label of the CIL setting.
In semantic segmentation tasks involving dense predictions, the problem of catastrophic forgetting typically becomes more challenging.
Most recent works~\cite{MiB,SSUL,PLOP} have struggled to alleviate this problem, and the freeze strategy~\cite{SSUL} (\ie, freezing most of the parameters during all the incremental steps after learning base classes) was shown to be the most efficient way in the state-of-the-art CISS methods~\cite{SSUL,MicroSeg}.

However, while the freeze strategy effectively alleviates catastrophic forgetting, it is a compromise for the model's plasticity, meaning the model becomes hard to adapt to new classes.
Furthermore, when considering the \textit{lifelong learning} setting with an infinite number of novel classes, the plasticity of the incremental learner will be crucial. 
Therefore, ideally, the optimal solution should be fine-tuning all the parameters for new classes, while catastrophic forgetting needs to be handled properly, rather than a freeze strategy.

Thus, to address the above-mentioned issues in CISS, especially the limitations of the freeze strategy, we prioritize the model's plasticity by a flexible parameters tuning strategy, and pursue the more discriminative feature representation for the balance to stability.
An intuitive idea for discriminative representation is to compute prototypes~(category centroids) for each category and apply contrastive learning to improve the diversity among categories~\cite{SDR,UCD}.
However, as mentioned in the Gaussian Mixture Model~(GMM)~\cite{reynolds2009gaussian}, the natural samples from the same category should come from the mixture of multiple Gaussian distributions.
Furthermore, several prior works~\cite{permuter2006study,pernkopf2005genetic,reynolds2009gaussian,PMM} have also claimed that the representation of categories in the feature space should be multiple activations, as shown in Fig.~\ref{figure:PMM}.

To this end, we propose the \textbf{Co}ntrast \textbf{in}ter- and \textbf{in}tra-class representations for Incremental \textbf{Seg}mentation~(\textbf{CoinSeg}), by pursuing discriminative representations.
Although the idea of adapting contrastive learning to CISS is intuitive, it is worth studying and critical to choose the appropriate areas, to contrast inter- and intra-class representations for incremental segmentation.
Firstly, the CoinSeg emphasizes intra-class diversity with multiple contrastive representation centroids.
Specifically, we identify regions with strong objectness using mask proposals, which are more likely to be instances/centroids of a particular category, as shown in Fig.~\ref{figure:PMM}~(b). 
We contrast these regional objectness to reinforce intra-class diversity and robust representation of the model.
In order to mitigate the potential bias from intra-class diversity, we incorporated category-level pseudo-labels to augment category-level consistency and inter-category diversity.
Meanwhile, we apply Swin Transformer~\cite{swintransformer} to better extract the local representation of the samples.
Additionally, CoinSeg ensures the model's stability and alleviates forgetting through a specific Flexible Tuning strategy.
In this strategy, a better balance between plasticity and stability is achieved by designing an initial learning rate schedule and regularization terms.

Finally, the CoinSeg outperforms prior methods in multiple benchmarks, especially in realistic and hard long-term scenarios VOC 10-1 and 2-2,
where our approaches show significant performance gains of 6.7\% and 17.8\%, comparing with previous state-of-the-art, respectively.

%% file: figures/fig1_PMM.tex
\begin{figure}[t]
\centering 
{\includegraphics[width=1.00\linewidth]{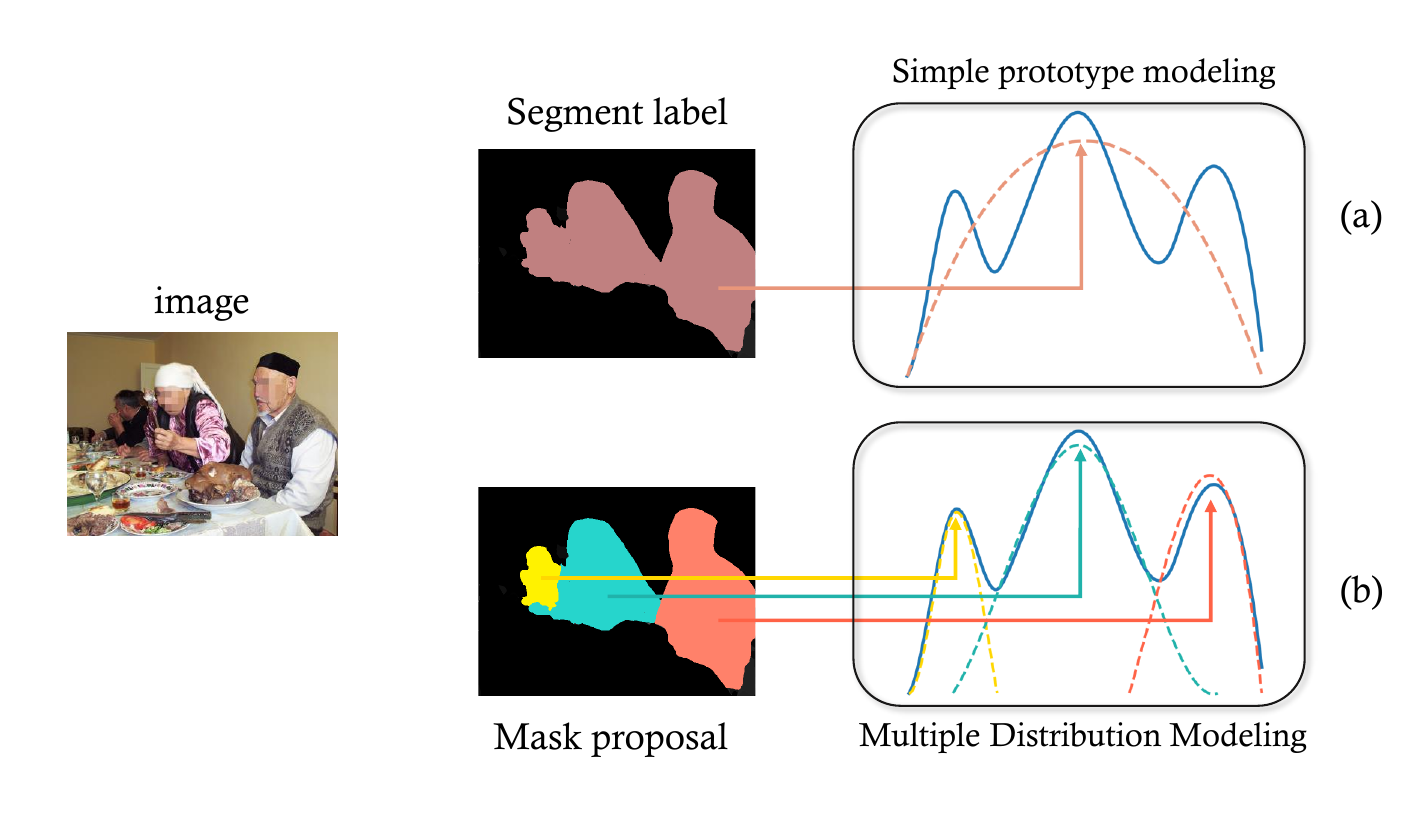}
}
\vspace{-0.5cm}
\caption{{Comparison of Simple Prototype Modeling and Multiple Distribution Modeling. (a): Simple prototype modeling to represent a category with global average pooling; (b): Multiple Gussian distribution modeling to represent a category with its regional objectness, which is discovered with the guidance of mask proposals. }}
\vspace{-0.3cm}
\label{figure:PMM}
 \end{figure}

%% file: sections/2_related_work.tex
\section{Related work}
\subsection{Class Incremental Learning} 
Class incremental learning~(CIL) is incremental learning focusing on defying catastrophic forgetting in classification tasks.
Replay-based approaches were proposed to store a sampler of historical data~\cite{ COPE_de2021continual,GEM_lopez2017gradient, ICaRL_rebuffi2017icarl,ER_rolnick2019experience,CLD_shin2017continual}, which can be used for future training to prevent forgetting.
The historical data can also be obtained with web search engine~\cite{RECALL} or generation models~\cite{ostapenko2019learning,shin2017continual,wu2018memory}.
Another intuitive thought to tackle CIL task is based on parameter isolation~\cite{EXPERT_aljundi2017expert,PACKNET_mallya2018packnet, DAN_rosenfeld2018incremental, PNN_rusu2016progressive, HAT_serra2018overcoming, zhang2023slca}.
Parameter isolation methods assign dedicated model parameters for each specific task, while bringing the continually increasing number of parameters with task increases. 
Regularization-based methods, such as knowledge distillation~\cite{MAS_aljundi2018memory,KD_hinton2015distilling,LFL_jung2016less, zhu2023ctp, han2023global} and restricting model training with networks trained at previous tasks~\cite{LwF_li2017learning,EBLL_rannen2017encoder,DMC_zhang2020class}, are also effective in tackling the catastrophic forgetting problem in incremental learning. These methods allow the model to transfer knowledge from previous tasks to new tasks, which can help prevent forgetting.

\subsection{Class Incremental Semantic Segmentation} 
Recently, there is a growing interest in the field of incremental learning for semantic segmentation(\ie, Class incremental semantic segmentation, CISS), and researchers are proposing various approaches to tackle CISS.
Modeling-the-Background~(MiB)~\cite{MiB} first remodeling the background~(dummy label) in the ground truth, and designs a distillation-based framework in CISS.
Douillard et al.~\cite{PLOP} proposed the approach of PLOP to define a pseudo label for CISS and proposes a local distillation method as an extended constraint based on MiB.
The SSUL~\cite{SSUL} first introduces replay-based approaches to the CISS task and maintains a memory bank of historical samples for future model training. Besides, the SSUL prevents the model from forgetting by freezing model parameters.
RCIL~\cite{RCIL} proposes a dual-branch architecture, in which one is freeze and the other is trainable, and introduces a channel-wise feature distillation loss.
MicroSeg~\cite{MicroSeg} introduces mask proposals to CISS and further clarifies image labels to tackle background shifts.

\subsection{Vision Transformer}
Transformers for computer vision (\ie, Vision Transformer) have attracted more and more attention. 
ViT~\cite{VIT} is the first widely known vision transformer, which transposes transformer directly to image classification tasks, achieving comparable performances with CNN-based methods. 
Since then, researchers have explored modifications and optimizations to ViT, proposing numerous designs~\cite{chu2021conditional,touvron2021training} for Vision Transformers that differ from those used in NLP.
Swin Transformer~\cite{swintransformer} proposes shifted window-based self-attention, bringing greater efficiency for vision transformer.
Other works have explored the use of transformers for image segmentation.
Segformer~\cite{xie2021segformer} consists of  hierarchically structured transformer layers to deal with semantic segmentation.
Mask2Former~\cite{mask2former} and MaskFormer~\cite{maskformer} propose universal transformer-based architectures for semantic, instance and panoptic segmentation.

%% file: sections/3_method.tex
\section{Method}

\subsection{Task Definition}
\label{sec:taskdef}
We define the task of Class Incremental Semantic Segmentation~(CISS) according to the common definition in previous works~\cite{MiB,SSUL,PLOP,SDR,RCIL}.
CISS is composed of a series of incremental \textit{learning steps}, as $t = 1,\dots,T$.
Each learning step has a sub-dataset $\bm{\mathcal{D}}^t$ and a corresponding class set of samples $\bm{\mathcal{C}}^t$. 
For any pair $(\bm x^{t},\bm{y}^{t})$ within $\bm{\mathcal{D}}^t$, $\bm x^{t}$ and $\bm{y}^{t}$ denote the input image and its ground-truth mask, respectively. 
We follow the definition of classes set learned in CISS as in MicroSeg~\cite{MicroSeg}:
In each learning step of class incremental semantic segmentation, the current classes set can be represented as the union of the classes to be learned, denoted by $\bm{\mathcal{C}}^{t}$, and a special class $c_{u}$ denotes ``areas does not belong to current foreground classes''.
From the perspective of each current learning step, the class $c_{u}$ can be interpreted as the background class, and its composition varies across different learning steps. 

The goal of the CISS model $f_{t,\bm \theta}$ with parameters $\bm \theta$ at the $t_{th}$ learning step is to assign a probability to each class for every pixel in $\bm x^{t}$.
The CISS model 
$f_{t,\bm\theta}$ is a composite of a feature extractor $g_{t,\bm\theta_1}$ and a classifier $h_{t,\bm\theta_2} $, which classifies each category in the union of $\bm{\mathcal{C}}^{1:t}= {\textstyle \bigcup_{i=1}^{t} \bm{\mathcal{C}}^i}$ and the unseen class $c_u$ (i.e., $\mathbb{C}^{t}=\bm{\mathcal{C}}^{1:t} \cup c_u$).
After the $t$-th learning step, the model also needs to provide a prediction for \textit{all seen classes} $\mathbb{C}^{t}$.
The prediction of CISS with model $f_{t,\bm\theta}$ can be expressed as
$
\hat{\bm y}_t=\arg\max_{c\in \mathbb{C}^{t}} f_{t,\bm\theta}^c(\bm x)
$.

\subsection{Contrast inter- and intra-class Representations}
\input{figures/diversity}
Previous studies~\cite{SSUL, MicroSeg} have highlighted the advantages of the freeze strategy in CISS. 
However, while parameter freezing can be beneficial, it may also restrict the model's plasticity and block further exploration of CISS. 
Unlike previous methods, our approach prioritizes model plasticity and allows fine-tuning to address this concern.
With the application of fine-tuning, it is possible to train the model to obtain more robust and discriminative representations.
Thus, we designed two contrastive losses for enhancing the ability of model representation learning to improve the model plasticity, \ie, pseudo label-guided representation learning for inter-class diversity and mask proposal-guided representation learning for intra-class diversity, as depicted in Fig.~\ref{figure:diversity}.
\subsubsection{Contrast inter-class representations}
We apply pseudo-label-guided representation learning to enhance inter-class diversity.
The approach also highlights category-level consistency to avoid bias from intra-class diversity, which introduces in Sec.~\ref{sec:intra_class}.
\paragraph{Pseudo labels.}
Due to the limitation of data acquisition in CISS, the current ground truth $\bm{y}^t$ is only annotated with classes at the current learning step, \ie, $\bm{\mathcal{C}}^t$.
So we need to extend the available ground truth $\bm{y}^t$ to include predictions from $f_{t-1}$, creating a more informative label for contrastive learning.
Specifically, we get the label prediction $\bm{\hat{y}}^{t-1} \in \mathbb{R}^{ |\bm{\mathcal{C}}^{1:t-1} \cup c_u| \times H \times W}$ with the inference of $f_{t-1}(\bm x)$ and the confidence $\bm{{s}}^{t-1} \in \mathbb{R}^{H\times W}$ of prediction. 
Formally:
\begin{equation}
\begin{aligned}
    \bm{\hat{y}}^{t-1}=\arg\max_{\bm{\mathcal{C}}^{1:t-1} \cup c_u} f_{t-1}(\bm x),\\
 \bm{{s}}^{t-1}=\max_{\bm{\mathcal{C}}^{1:t-1} \cup c_u} \sigma( f_{t-1}(\bm x)),
\end{aligned}
\end{equation}
where $\sigma(\cdot)$ denotes Sigmoid function. 
With the ground truth mask $\bm{y}^{t}=\{\bm{y}^t_i\}$ in current step, we mix the supervision label $\tilde{\bm{y}}^t_i$ of pixel $i$ according to the following rules: 

\begin{equation}
\label{eq:modelbg}
\resizebox{0.9\columnwidth}{!}{$
\begin{aligned}
    \tilde{\bm{y}}^t_i =
    \begin{cases}
    {\bm{y}}^t_i & \text{ where }
    {\bm{y}}^t_i\in \bm{\mathcal{C}}^t \text{ or } \bm{y}^t_i= c_{u} \wedge \bm{{s}}^{t-1}_i<\tau \\
    \bm{\hat{y}}^{t-1}_i & \text{ where }\bm{y}^t_i= c_{u} \wedge \bm{{s}}^{t-1}_i \ge \tau \\
    \end{cases},
    \end{aligned}$
    }
\end{equation}
    
where threshold $\tau=0.7$, `$\wedge$' represents the co-taking of conditions.
\paragraph{Inter-class contrastive loss.}
With the guidance of pseudo label, CoinSeg gets the prototypes    of classes~(\ie, class centroids), and sets up contrastive loss to better represent inter-class diversity.

Given the feature maps $\bm{M}^{t}=g_t(\bm x)$ and binary masks $\tilde{\bm{y}}^t \in \{0,1\}^{|\mathbb{C}^{t}|\times h \times w}$ from the pseudo labels, CoinSeg applies masked average pooling~(MAP)~\cite{SGONE_zhang2020sg} to obtain prototypes of each foreground class as $\bm {P}_{int}^{t}$ for contrast inter-class representation:

\begin{equation}
\resizebox{0.9\columnwidth}{!}{
    $\bm {P}_{int}^{t} = MAP(\bm{M}^{t}, \tilde{\bm{y}}^{t})=\frac{\sum_{i=1, j=1}^{h, w} (\tilde{\bm{y}}^{t,i,j} * {M}^{t,i, j})}{\sum_{i=1, j=1}^{h, w} \tilde{\bm{y}}^{t,i,j}}$.
}
\end{equation}

We abbreviate the operation as $\bm {P}_{int}^{t} = MAP (\bm{M}^{t}, \tilde{\bm{y}}^{t})$. $\bm {P}_{int}^{t}$ is a series of vectors as prototypes, \ie, discriminative representation of each classes.
At learning step $t>1$, CoinSeg adapts model $f_{t-1}$ from learning step $t-1$ as guidance to train the current model $f_t$, and we note $\bm{M}^{t-1}=g_{t-1}(\bm x)$.
$\bm {P}^{t-1}_{int}$ can also be obtained with feature maps of previous step $\bm{M}^{t-1}$ through a similar operation.
CoinSeg assigns prototypes clustered from the same pseudo label as positive pairs in contrastive learning, and prototypes from different labels as negative pairs. 

The distance of prototypes is measured with the inner product $Dot(\cdot,\cdot)$, and the contrastive loss with the guidance of the pseudo label can be expressed as: 
\begin{equation}
\resizebox{0.9\columnwidth}{!}{$
    \begin{aligned}
    \mathcal{L}_{int}=-\frac{1}{K}\sum_{i=1}^{K}\log\frac{\exp(Dot(\bm{P}^{t,i}_{int}, \bm{P}^{t-1,i}_{int}))}{\sum_{j=1,j\neq i}^{2K}\exp(Dot(\bm{P}^{t,i}_{int},\bm{P}^{*,j}_{int}))},
    \end{aligned}$
}
\end{equation}
where $\bm{P}^{*}_{int} =\bm{P}^{t}_{int}\cup\bm{P}^{t-1}_{int}$, $K=|\bm{P}^t_{int}|$ is the number of inter-class prototypes, thus $2K=|\bm{P}^t_{int}|+|\bm{P}^{t-1}_{int}|$ .
\input{figures/contrastive}
Generally, as shown in Fig.~\ref{figure:contrastive}, we abbreviate the contrastive loss by previous operations with the guidance of mask $\tilde{\bm{y}}^{t}$ as:
\begin{equation}
    \mathcal{L}_{int} = CON(\tilde{\bm{y}}^{t}, M^{t}, M^{t-1} ).
\end{equation}

\subsubsection{Contrast intra-class representations}
\label{sec:intra_class}

Inspired by the Gaussian mixture models, we focus on intra-class diversity benefits to more robust representation learning. 
Thus we mine potential regional objectness within categories, and emphasize intra-class diversity through mask proposals-guided representation learning.
\paragraph{Mask proposals.} 
CoinSeg adapts a set of \textit{class-agnostic} binary mask $\bm{B}\in\{0,1\}^{N\times H\times W}$ as proposals~(\ie, mask proposals), where $N$ denotes the number of mask proposals. 
Following the practice of~\cite{MicroSeg}, mask proposals are generated with Mask2Former~\cite{mask2former}.
Note each pixel in an image belongs and only belongs to one of the mask proposals.

\paragraph{Intra-class contrastive loss.}
Mask proposals discover regional objectness in images, which are likely to be diverse instances or centroids of a category, and benefit the construction of intra-class contrastive learning.
For $\bm M^{t}=g_t(\bm x)$, CoinSeg obtains prototypes of each mask proposal by $\bm{P}^{t}_{itr}=MAP(\bm{M}^t$, $\bm{B})$. 
Namely, $\bm{P}^{t}_{itr}$ is a series of mask proposal-based prototypes with size of $N\times C$.
$\bm{P}^{t-1}_{itr}$ can also be obtained with the similar operation.
To better characterize the intra-class diversity by contrastive learning, CoinSeg assigns prototypes from the same mask proposal as positive pairs, and prototypes from different mask proposals as negative pairs.
The contrastive loss with the guidance of mask proposals can be expressed as:
\begin{equation}
    \resizebox{0.9\columnwidth}{!}{$
    \begin{aligned}
    \mathcal{L}_{itr} &= CON(\bm{B}, \bm{M}^{t}, \bm{M}^{t-1})\\
    &=-\frac{1}{N}\sum_{i=1}^{N}\log\frac{\exp(Dot(\bm{P}^{t,i}_{itr}, \bm{P}^{t-1,i}_{itr}))}{\sum_{j=1,j\neq i}^{2N}\exp(Dot(\bm{P}^{t,i}_{itr},\bm{P}^{*,j}_{itr}))}.
    \end{aligned}$
    }
\end{equation}
\subsubsection{Summary}

As explained in Sec.~\ref{sec:taskdef}, due to the incremental learning task, there is a limitation in the acquisition of category labels in the ground truth, which leads to the fact that it is difficult for the CISS model to distinguish between categories.  To this end, we enhance the model's category discrimination ability by emphasizing inter- and intra-class diversity. Moreover, contrastive learning overcomes catastrophic forgetting by the design of positive contrast pairs, \ie, constraining the corresponding prototypes from feature map $\bm{M}^{t}$ and $\bm{M}^{t-1}$ to be consistent. This knowledge distillation-like mechanism helps to alleviate forgetting and improve performance in CISS.

Finally, the total loss for the learning of the Contrast inter- and intra-class Representations is:
\begin{equation}
    \begin{aligned}
    \mathcal{L}_{ct} =  \mathcal{L}_{int} + \mathcal{L}_{itr}.
    \end{aligned}
\end{equation}

\subsection{Flexible Tuning Strategy}
As mentioned before, releasing parameter training for plasticity could lead to the best performance, but more discriminative feature representation is needed.
Although we have designed the Contrast inter- and intra-class Representations for the discriminative representation, some more specific parameter tuning strategy is necessary to ensure stability~(\ie, handling catastrophic forgetting) as well.
Therefore, we introduce the Flexible Tuning~(FT) strategy as that, which allows for training the model while mitigating the effects of forgetting, achieving a balance between stability and plasticity.
\input{figures/Flexible_freeze_overview.tex}
Fig.~\ref{figure:FF} shows the comparison of the freeze strategy and flexible tuning strategy.
The freeze strategy involves keeping the parameters of the feature extractor and classifier for historical classes fixed for learning step $t>1$.
In contrast, the flexible tuning strategy uses a lower learning rate and regularization constraints to allow for more flexible adjustments of these parameters.
\paragraph{Flexible initial learning rate.}
The segmentation model $f_{\bm \theta}$ is represented with a feature extractor $g_{\bm \theta_1}$ and a classifier $h_{\bm \theta_2}$, with parameters set $\bm \theta_1$ and $\bm \theta_2$ accordingly.
Unlike the freeze strategy~\cite{SSUL}, which sets the learning rate to zero for both $\bm \theta_1$ and $\bm \theta_2$ of historical classes, 
the FT strategy applies a \textbf{initial} learning rate schedule of each learning step to these parameters.
Specifically, during the first learning step~($t=1$), the initial learning rate $lr$ will be kept as original $lr_{0}$.
For subsequent learning steps~($t>1$), the initial learning rate is gradually reduced to better preserve the knowledge of the historical categories.
So the initial $lr$ of step $t$ is expressed as:
\begin{equation}    
    \label{eq:lrschedule}
    lr^t =
    \begin{cases}
     lr_{0} & \text{ if } t=1 \\
     e^{-t} \lambda_{lr} \cdot lr_{0} & \text{ if } t>1 
    \end{cases},
\end{equation}
 where $\lambda_{lr}$ is a hyper-parameter, and $e^{-t}$ refers to the exponential decay of the learning rate.
\paragraph{Regularization constraints.}
When the flexible learning schedule allows the model to adapt more to new concepts, regularization constraints for mitigating forgetting become even more crucial than before.
To better alleviate forgetting and ensure stability when the model adapts to new categories, CoinSeg applies some regularization constraints.
To be more specific, for a sample $\bm x$, CoinSeg extract feature map $\bm{M}^{t-1}\in\mathbb{R}^{C\times H'\times W'}$ by the feature extractor $g_{t-1}$. \ie, $\bm{M}^{t-1}=g_{t-1}(\bm x)$, and 
$C$ denotes the number of channels of $\bm{M}_{t-1}$. 
Similarly, $\bm{M}^t = g_t(\bm{x})$. Then, CoinSeg constrains the consistency of $\bm{M}^t$ and $\bm{M}^{t-1}$ with the Mean Square Error~(MSE) as:
\begin{equation}
    \label{LFKD}
    \begin{aligned}
    \mathcal{L}^{F}_{kd} &= MSE(\bm{M}^{t},\bm{M}^{t-1}) \\
    &= \frac{1}{C} \frac{1}{H'W'} \sum_{j=1}^{C} \sum_{i=1}^{H'W'} (\bm{M}^{t}_{i,j}-\bm{M}^{t-1}_{i,j})^2.
    \end{aligned}
\end{equation}

Furthermore, CoinSeg also adapts knowledge distillation of logits $z^t$ and $z^{t-1}$. 
Logits is the output of the model, \ie, $z^t = f_t(x)$. 
Note $z^t \in \mathbb{R}^{ |\bm{\mathcal{C}}^{1:t} \cup c_u |\times H \times W}$, where $\bm{\mathcal{C}}^{1:t}= {\textstyle \bigcup_{i=1}^{t} \bm{\mathcal{C}}^i}$ is all seen categories of step $t$, including historical and present classes. 
As for logits $z^{t-1} \in \mathbb{R}^{ |\bm{\mathcal{C}}^{1:t-1}\cup c_u|\times H\times W}$, while $\bm{\mathcal{C}}^t$ is not exist for previous model $f_{t-1}$. 
Following the common practice~\cite{MiB,PLOP,SDR,RCIL}, CoinSeg adopts the following approaches to remodeling the logits $\hat{z}^t$ for pixel $i$ with class $c$ :
\begin{equation}    
    \label{eq:MiB}
    \hat{z}^t_{i,c} =
    \begin{cases}
     {z}^t_{i,c} & \text{ if } c \ne c_u \\
     \sum_{j \in \bm{\mathcal{C}}^t} {z}^t_{i,j} & \text{ if }c = c_u 
    \end{cases}.
\end{equation}
Then cross entropy loss $L^{z}_{kd}$ is adapted to distill logits:
\begin{equation}
    \label{LzKD}
    \begin{aligned}
    \mathcal{L}^{z}_{kd} &= CE(\hat{z}^t,z^{t-1}) \\
    &= -\frac{1}{|\mathbb{C}^{t-1}|} \frac{1}{H'W'} \sum_{j=1}^{|\mathbb{C}^{t-1}| } \sum_{i=1}^{ H'W'} (\hat {z}^{t}_{i,j}\log z^{t-1}_{i,j}),
    \end{aligned},
\end{equation}
where we assign $|\mathbb{C}^{t-1}| = |\bm{\mathcal{C}}^{1:t-1}\cup c_u| $.

In summary, the regularization constraints in the flexible tuning strategy are:
\begin{equation}
    \label{Lreg}
    \begin{aligned}
    \mathcal{L}_{reg} =  \mathcal{L}^{F}_{kd} + \mathcal{L}^{z}_{kd}
    \end{aligned}.
\end{equation}

\subsection{Objective Function}
Following the practice of previous methods\cite{MicroSeg}, CoinSeg adapt common Binary Cross-Entropy~(BCE) loss as supervised segmentation loss $\mathcal{L}_{BCE}$ with the augmented label $\tilde{\bm{y}}$,
and the final objective function of CoinSeg is:
\begin{equation}
   \mathcal L = \mathcal{L}_{BCE} +\lambda_{c}\cdot \mathcal{L}_{ct} + \lambda_{r}\cdot \mathcal{L}_{reg}.
\end{equation}

%% file: figures/diversity.tex
\begin{figure}[t]
\centering 
{\includegraphics[width=1.00\linewidth]{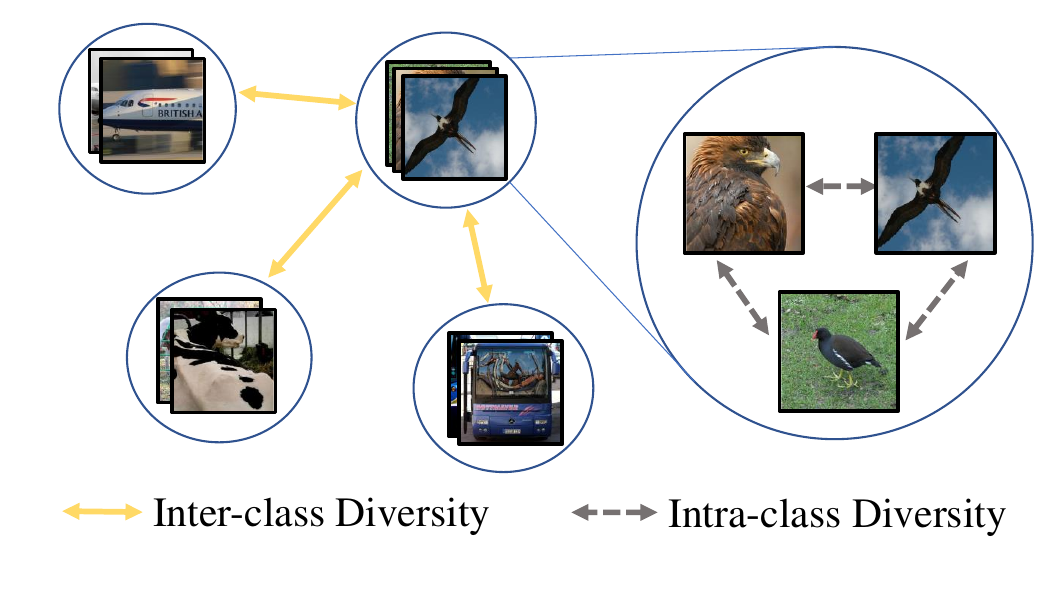}
}
\vspace{-0.5cm}
\caption{Illustration of the inter- \& intra-class diversity. Only the contrastive learning with class ``bird'' is shown for example.}
\vspace{-0.4cm}
\label{figure:diversity}
 \end{figure}

%% file: figures/contrastive.tex
\begin{figure}[t]
\centering 
{\includegraphics[width=1.00\linewidth]{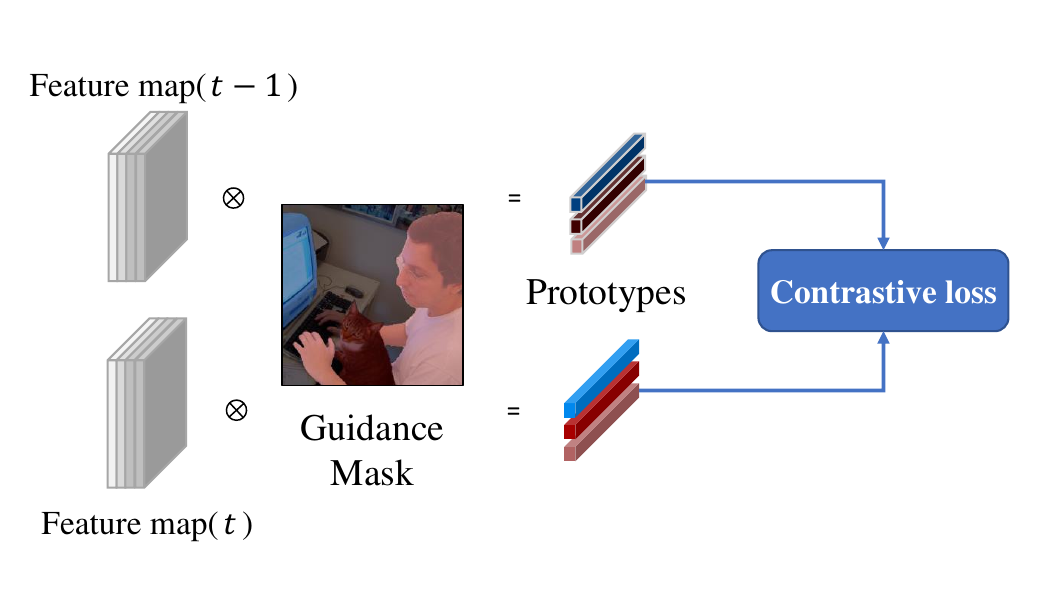}
}
\vspace{-0.5cm}
\caption{{The pipeline of contrastive learning.}}
\vspace{-0.3cm}
\label{figure:contrastive}
 \end{figure}

%% file: figures/Flexible_freeze_overview.tex
\begin{figure}[t]
\centering 
{\includegraphics[width=1.00\linewidth]{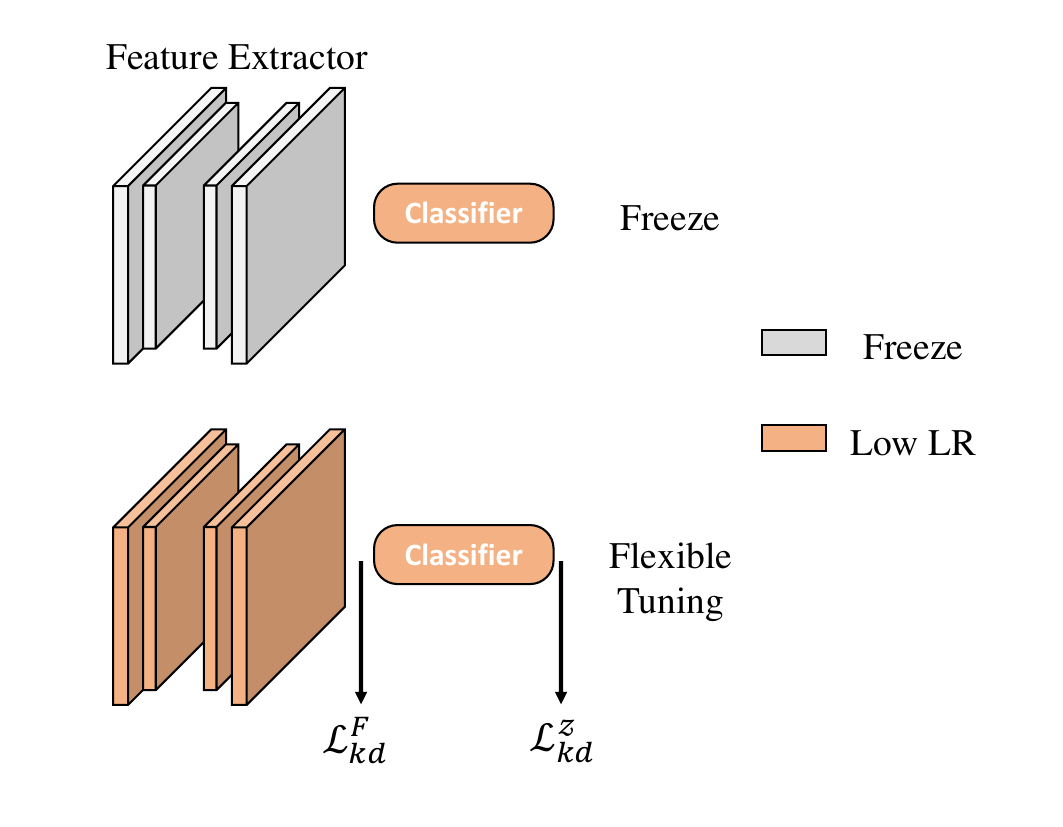}
}
\vspace{-0.5cm}
\caption{{Comparison of Freeze and Flexible Tuning strategy for learning step $t>1$. Best viewed in color.}}
\vspace{-0.3cm}
\label{figure:FF}
 \end{figure}

%% file: sections/4_experiment.tex
\section{Experiment}
\subsection{Experimental Setups}
\paragraph{Dataset.}
We have evaluated our approach on datasets of Pascal VOC 2012~\cite{VOC_everingham2010pascal} and ADE20K~\cite{ADE_zhou2017scene}.
Pascal VOC 2012 contains 10,582 training images and 1449 validation images, with a total of 20 foreground classes and one background class.
ADE20K contains 20,210 training images and 2,000 validation images with 100 thing classes and 50 stuff classes. 
\paragraph{Protocols.}
Following the conventions of previous work \cite{SSUL,PLOP,MicroSeg}, we mainly evaluate our approach for CISS with \textit{overlapped} experimental setup, which is more realistic than the \textit{disjoint} setting that was studied by \cite{MiB,SDR,RCIL}.
For each benchmark dataset, we examine our approach under multiple \textit{incremental scenarios}.
We abbreviate each incremental scenario in the form of $X-Y$, 
where $X$ means the initial number of base classes, and $Y$ refers to the incremental number of classes in each step.
For instance, the 10-1 scenario of the VOC dataset~(VOC 10-1) signifies 10 classes learned at the first learning step, and
 1 incremental class learned in each subsequent step, \ie, VOC 10-1 scenario takes 
a total of 11 steps to learn the entire dataset.

\input{tables/VOC}

\paragraph{Implementation details.}
In accordance with the established practice~\cite{MiB,SSUL,SDR}, we use DeepLabv3~\cite{chen2017deeplab} as the segmentation network.
CoinSeg chooses Swin Transformer-base~(Swin-B)~\cite{swintransformer} pretrained on ImageNet-1K as the backbone.
Swin Transformer provides a better feature representation of the local patches, which is concerned in our CoinSeg.
For the segmentation head, we apply the dual-head architecture from MicroSeg, including dense prediction branch and proposal classification branch~\cite{MicroSeg}.
We optimize the model by ADAMW~\cite{adamw} with an learning rate of $lr_{0} = 10^{-4}$. 
The batch size is 16 for Pascal VOC 2012, and 12 for ADE20K.
The window size for Swin Transformer is 12.
Data augmentation~\cite{SSUL} are applied for all samples.
To ensure a fair comparison with previous methods, we apply the same class-agnostic mask proposals with MicroSeg~\cite{MicroSeg}. 
Specifically, mask proposals are generated with parameters-fixed Mask2Former~\cite{mask2former} pre-trained on MS-COCO, 
with $N=100$ for all experiments. 
To prevent information leakage, the Mask2Former is \textbf{not} fine-tuned on any benchmark dataset~\cite{MicroSeg}.
All experiments are implemented with PyTorch on two NVIDIA GeForce RTX 3090 GPUs.
Hyper-parameters $\lambda_{lr}=10^{-3}$, $\lambda_c=0.01$ and $\lambda_r=0.1$  are set for all experiments.

\paragraph{Baselines.} 

We evaluate our CoinSeg on multiple incremental scenarios. 
The performance of CoinSeg is compared with some representative approaches in CIL, including Lwf-MC~\cite{LwF_li2017learning} and
ILT~\cite{ILT_michieli2019incremental}, which are applied to the experimental setup of CISS.
Besides, we provide the results of the prior state-of-the-art CISS methods, including MiB~\cite{MiB}, SDR~\cite{SDR}, PLOP~\cite{PLOP}, SSUL~\cite{SSUL}, RCIL~\cite{RCIL}, and MicroSeg~\cite{MicroSeg}. 
To ensure a fair comparison with recent state-of-the-art CISS methods~\cite{SSUL,MicroSeg} that employ different backbones, we produced results by replacing their backbones with Swin-B.
These methods are re-implemented with their official codes. 
Methods with suffix `M'~(for example, \textit{SSUL-M}) are with memory sampling strategy~\cite{SSUL}, which maintains a memory bank of historical samples for rehearsal at future learning steps. 
Besides, we also provide the experimental results of \textit{joint} training~(training all classes together), on both Resnet101 and Swin-B backbones.
The result of joint training is usually regarded as the upper bound of incremental learning~\cite{MiB,LwF_li2017learning}, \ie, offline training.
The mean Intersection-over-Union (mIoU) is applied as the evaluation metric for all experiments and analyses.
For each method in CISS,  three perspectives are presented: the performance of base classes, novel classes, and total performance, respectively.
Please refer to the supplementary material for more details.

\subsection{Experimental Results}
\paragraph{Comparison on VOC.}

\input{figures/comparison}

On the Pascal VOC 2012 dataset, we evaluate CoinSeg on various incremental scenarios, including long-term scenarios with lots of learning steps~(10-1, 15-1), with a large number of base classes~(19-1, 15-5) and an equally-divided long-term scenario~(2-2). 
The performance comparison of our approach with classical CIL methods and prior CISS methods is presented in Tab.~\ref{tab:voc}.
Our CoinSeg exhibits a significant performance advantage in all incremental scenarios, even when compared to the state-of-the-art methods using Swin Transformer as the backbone.

In particular, in very long-term incremental scenarios with few base classes, like 10-1 and 2-2, our CoinSeg brings huge performance gaps of 6.7\% and 17.8\%, respectively, in comparison to MicroSeg.
The freeze strategy in SSUL and MicroSeg prevented these methods from performing well in incremental scenarios with few base classes. 
In such scenarios, the base classes contain fewer concepts, which can easily introduce bias for the representation learning of novel classes. 
Our approach addresses this challenge by allowing the model to adapt to novel classes with the design of `contrast inter- and intra-class representations'.
In such incremental scenarios, the base classes contain fewer concepts and can easily introduce bias for the representation learning of novel classes.
Besides, we concern the scenarios with a large number of base classes~(\ie, 19-1 and 15-5), in
which, CoinSeg achieves state-of-the-art with a performance gain of 3.4\% and 2.4\%, compared with MicroSeg.

\input{tables/ADE}

Besides, Fig.~\ref{fig:compareVOC}~(a,b) depicts the variations of mIoU of \textbf{all seen classes} by learning steps during incremental learning, in two long-term incremental scenarios, \ie, VOC 15-1 and VOC 2-2, respectively.
The performance of each step depends on two factors: 1) the forgetting of historical classes, and 2) the ability to learn new classes.
Fig.~\ref{fig:compareVOC}~(a) shows the results of VOC 15-1, which involves a large number of base classes. 
The performance of our CoinSeg exhibits the least decrease with increasing training steps, suggesting that CoinSeg causes less forgetting of previously learned knowledge than prior methods.
In contrast, while VOC 2-2 consists of 18 novel classes, results of Fig.~\ref{fig:compareVOC}~(b) clearly demonstrate that our CoinSeg is more adaptable to learning new classes, compared to previous methods. 
Additionally, our CoinSeg even shows a significant performance improvement while learning new classes in some steps (steps 3, 4, and 8).
Besides, to demonstrate the robustness of CoinSeg, we provide experimental results of VOC 15-1 with 20 different class incremental orders, including average and standard variance of mIoU, illustrate as Fig.~\ref{fig:compareVOC}~(c).
The results indicate that our approach performs better and is more robust, with a lower standard variance.

\input{tables/ourmethod_M}

\input{tables/detailed_ablation}

\paragraph{Comparison on ADE.}
For the more challenging ADE dataset, we choose incremental scenarios that have been widely compared in past methods, including 100-5, 100-10, 100-50, and 50-50, shown in Tab.~\ref{tab:ade}.
Our method, CoinSeg,  outperforms previous state-of-the-art once again, which shows that our method is not specific to a particular dataset and is effective under multiple benchmarks.

\subsection{Ablation Studies}

\paragraph{Overall ablation of CoinSeg.}
In this section, we present an evaluation of each specific design of the proposed CoinSeg, including the Swin Transformer, Flexible Tuning~(FT) strategy, and Contrast inter- and intra-class Representations~(\textit{Coin}).
As shown in Tab.~\ref{tab:detailed_abl}, we can find that all these proposed designs benefit the performance of CISS.
By replacing the backbone with Swin-Transformer~(row 1 \vs row 2), it improves the ability of feature representation of the feature extractor, which is especially beneficial for the representation of local features, and necessary for CoinSeg.
The comparison between row 2 and row 3 reflects the experiment of replacing the freeze strategy with the FT strategy, showing that the FT enhances model plasticity, allowing adaptation to novel classes, and slightly improving the performance.
The application of \textit{Coin} (row3 \vs row4) significantly improves the performance by imposing explicit constraints on intra-class and inter-class diversity, thereby enhancing the model's ability to represent data.
Overall, these components enable CoinSeg to achieve state-of-the-art performance.
\paragraph{Number of proposal in  $\mathcal{L}_{itr}$.}
We also investigated the influence of hyperparameter $N$ on the method's performance concerning \textit{contrast intra-class diversity}. Specifically, we explored how altering the number of proposals would affect the proposed CoinSeg. The results are presented in Tab.~\ref{tab:proposal_N_abl}. It can be observed that with $N$ increases, there is a slight improvement in performance, albeit with diminishing returns as the cardinality grows. Consequently, considering a trade-off between performance and computational complexity, we choose $N=100$ in our method.

\input{tables/ablation_proposal_N}

\subsection{Qualitative Analysis}
We have conducted a qualitative analysis using two examples in Fig.~\ref{figure:vis}.
In the first example~(rows 1,3 \& 5), during incremental learning, CoinSeg is able to retain knowledge about past classes and accurate predictions for them, whereas prior methods exhibit forgetting and misclassification, which demonstrated the stability of CoinSeg.
The second example~(rows 2, 4 \& 6), demonstrates the plasticity of our method, which refers to the ability to learn new classes.
For example, while prior methods predicted wrong bounds for the class `train' in incremental learning, our method is better adapted to these new classes making appropriate predictions.

\input{figures/visualization}

\subsection{Expansibility of CoinSeg}
\paragraph{Flexible tuning on prior methods.}

\input{tables/expandion}

As claimed, the FT strategy flexibly releases parameter training for plasticity and provides some more specific parameter tuning ways to ensure stability as well.
To better validate the effectiveness of the FT strategy, we applied the FT strategy to the state-of-the-art method MicroSeg, and compared the results to its original freeze strategy, as shown in Tab.~\ref{tab:abl_FL}.
The results demonstrate that on both backbones, applying the FT strategy leads to better performance, particularly, for the new classes, which proves that model plasticity is significantly enhanced by the FT strategy.
It means that the FT strategy can be also extended to other CISS methods for better performance, especially performance relying on model plasticity.

\paragraph{CoinSeg with memory sampling.}
The memory sampling strategy alleviates forgetting by rehearsing samples from past learning steps, and significantly improves performance~\cite{SSUL,MicroSeg}.
Thus, we produce the results of CoinSeg with this strategy, denoted as CoinSeg-M, for comparison with prior methods with the same strategy.
As shown in Tab.~\ref{tab:abl_M}, our CoinSeg~(the 5$_{th}$ row) achieves better performance in almost all incremental scenarios, even when compared with previous work using the sampling strategy.
When equipped with the memory sampling method, \ie, CoinSeg-M, it undoubtedly achieves state-of-the-art performance.

%% file: tables/VOC.tex
\begin{table*}[t!]
  \centering
\caption{
    Comparison with state-of-the-art methods on Pascal VOC 2012. \dag: Re-implemented with Swin-B backbone; \textcolor{lightgray}{Joint} is the upperbound.
  }
  \label{tab:voc}
  \begin{adjustbox}{max width=\linewidth}
  \begin{tabular}{l|c|ccc|ccc|ccc|ccc|ccc}
\toprule
\multirow{2}{*}{Method} & \multirow{2}{*}{Backbone} & \multicolumn{3}{c|}{\textbf{VOC 10-1 (11 steps)}} & \multicolumn{3}{c|}{\textbf{VOC 15-1 (6 steps)}} & \multicolumn{3}{c|}{\textbf{VOC 19-1 (2 steps)}} & \multicolumn{3}{c|}{\textbf{VOC 15-5 (2 steps)}} & \multicolumn{3}{c}{\textbf{VOC 2-2 (10 steps)}} \\
 &  & 0-10 & 11-20 & all & 0-15 & 16-20 & all & 0-19 & 20 & all & 0-15 & 16-20 & all & 0-2 & 3-20 & all \\ \midrule
\textcolor{lightgray}{Joint} & \textcolor{lightgray}{Resnet101} & \textcolor{lightgray}{82.1}  & \textcolor{lightgray}{79.6} & \textcolor{lightgray}{80.9} & \textcolor{lightgray}{82.7} & \textcolor{lightgray}{75.0} & \textcolor{lightgray}{80.9} & \textcolor{lightgray}{81.0} & \textcolor{lightgray}{79.1} & \textcolor{lightgray}{80.9} & \textcolor{lightgray}{82.7} & \textcolor{lightgray}{75.0} & \textcolor{lightgray}{80.9} & \textcolor{lightgray}{76.5} & \textcolor{lightgray}{81.6} & \textcolor{lightgray}{80.9} \\
LwF-MC~\cite{LwF_li2017learning} & Resnet101 & 4.7 & 5.9 & 4.9 & 6.4 & 8.4 & 6.9 & 64.4 & 13.3 & 61.9 & 58.1 & 35.0 & 52.3 & 3.5 & 4.7 & 4.5\\
ILT~\cite{ILT_michieli2019incremental} & Resnet101 & 7.2 & 3.7 & 5.5 & 8.8 & 8.0 & 8.6 & 67.8 & 10.9 & 65.1 & 67.1 & 39.2 & 60.5 & 5.8 & 5.0 & 5.1 \\
MiB~\cite{MiB} & Resnet101 & 12.3 & 13.1 & 12.7 & 34.2 & 13.5 & 29.3 & 71.4 & 23.6 & 69.2 & 76.4 & 50.0 & 70.1 & 41.1 & 23.4 & 25.9 \\
SDR~\cite{SDR} & Resnet101 & 32.1 & 17.0 & 24.9 & 44.7 & 21.8 & 39.2 & 69.1 & 32.6 & 67.4 & 57.4 & 52.6 & 69.9 & 13.0 & 5.1 & 6.2 \\
PLOP~\cite{PLOP} & Resnet101 & 44.0 & 15.5 & 30.5 & 65.1 & 21.1 & 54.6 & 75.4 & 37.4 & 73.5 & 75.7 & 51.7 & 70.1 & 24.1 & 11.9 & 13.7 \\
RCIL~\cite{RCIL} & Resnet101 & 55.4 & 15.1 & 34.3 & 70.6 & 23.7 & 59.4 & 68.5 & 12.1 & 65.8 & 78.8 & 52.0 & 72.4 & 28.3 & 19.0 & 19.4 \\
SSUL~\cite{SSUL} & Resnet101 & 71.3 & 46.0 & 59.3 & 77.3 & 36.6 & 67.6 & 77.7 & 29.7 & 75.4 & 77.8 & 50.1 & 71.2 & 62.4 & 42.5 & 45.3 \\
MicroSeg~\cite{MicroSeg} & Resnet101 & 72.6 & 48.7 & 61.2 & 80.1 & 36.8 & 69.8 & 78.8 & 14.0 & 75.7 & 80.4 & 52.8 & 73.8 & 61.4 & 40.6 & 43.5 \\
\midrule
\textcolor{lightgray}{Joint\dag} & \textcolor{lightgray}{Swin-B} & \textcolor{lightgray}{82.4} & \textcolor{lightgray}{83.0} & \textcolor{lightgray}{82.7} & \textcolor{lightgray}{83.8} & \textcolor{lightgray}{79.3} & \textcolor{lightgray}{82.7} & \textcolor{lightgray}{82.6} & \textcolor{lightgray}{84.4} & \textcolor{lightgray}{82.7} & \textcolor{lightgray}{83.8} & \textcolor{lightgray}{79.3} & \textcolor{lightgray}{82.7} & \textcolor{lightgray}{75.8} & \textcolor{lightgray}{83.9} & \textcolor{lightgray}{82.7} \\
SSUL\dag\cite{SSUL} & Swin-B & 74.3 & 51.0 & 63.2 & 78.1 & 33.4 & 67.5 & 80.8 & 31.5 & 78.4 & 79.7 & 55.3 & 73.9 & 60.3 & 40.6 & 44.0 \\
MicroSeg\dag~\cite{MicroSeg} & Swin-B & 73.5 & 53.0 & 63.8 & 80.5 & 40.8 & 71.0 & 79.0 & 25.3 & 76.4 & 81.9 & 54.0 & 75.2 & 64.8 & 43.4 & 46.5 \\
\midrule
\methodname~(Ours) & Swin-B & \textbf{80.1} & \textbf{60.0} & \textbf{70.5} & \textbf{82.7} & \textbf{52.5} & \textbf{75.5} & \textbf{81.5} & \textbf{44.8} & \textbf{79.8} & \textbf{82.1} & \textbf{63.2} & \textbf{77.6} & \textbf{70.1} & \textbf{63.3} & \textbf{64.3} \\ 
\bottomrule
\end{tabular}
  \end{adjustbox}
\end{table*}

%% file: figures/comparison.tex
 \begin{figure*}[t]
    \vspace{-5mm}
    \centering
    \subfigure[]{           
        \includegraphics[width=0.30\linewidth]{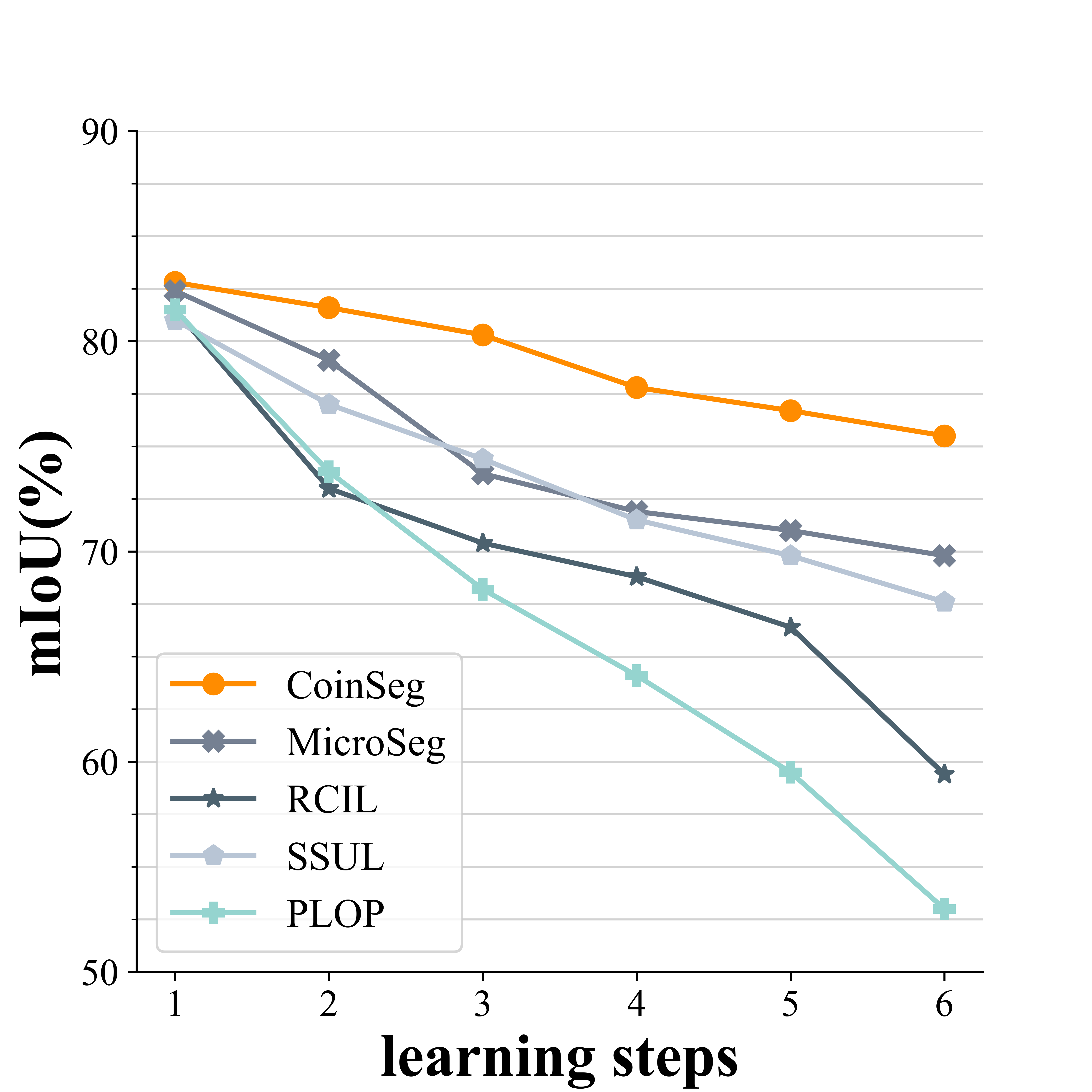} 
        \label{fig:miou_evo}  
    }
    \subfigure[]{           
        \includegraphics[width=0.30\linewidth]{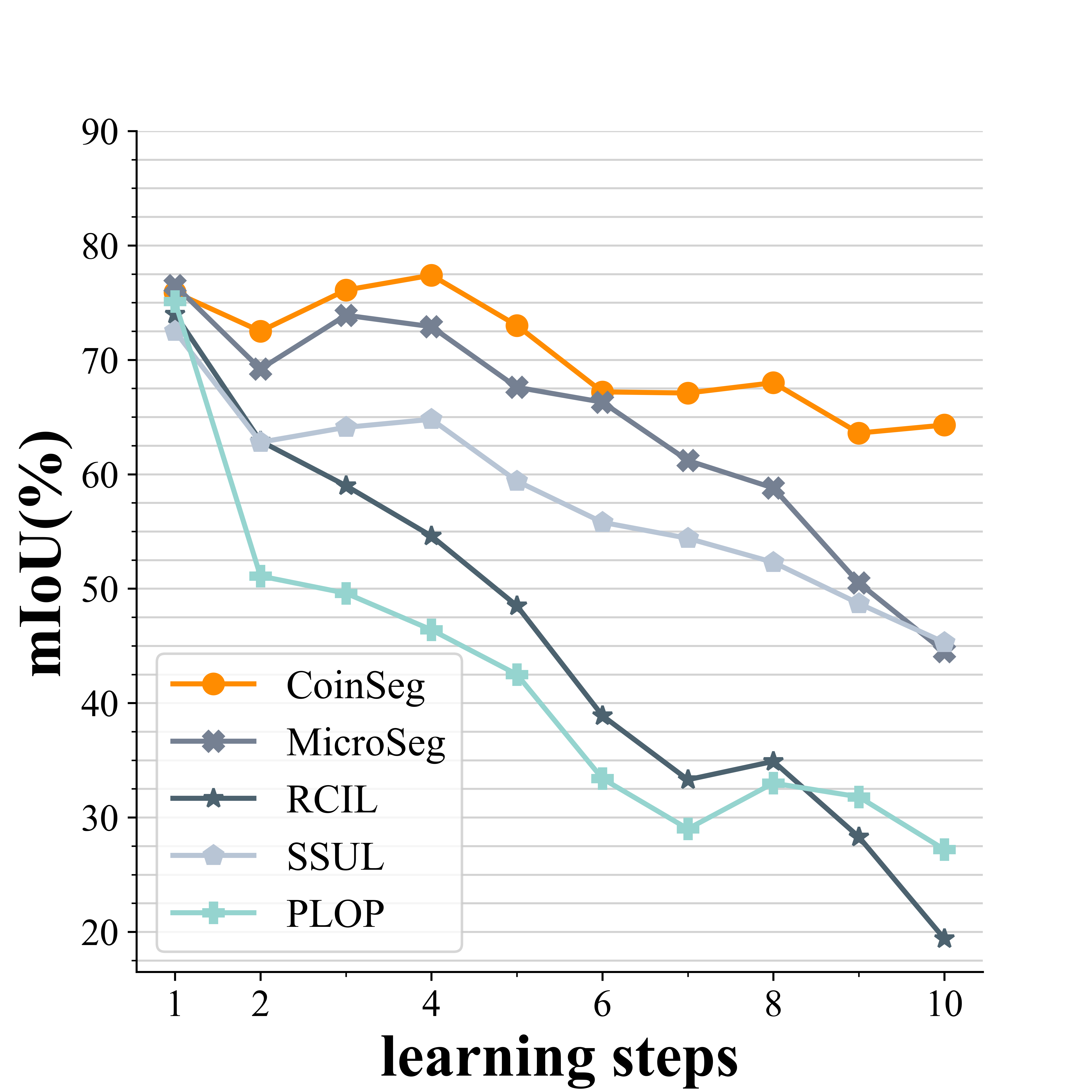} 
        \label{fig:class_ordering}         
    }
    \subfigure[]{           
        \includegraphics[width=0.30\linewidth]{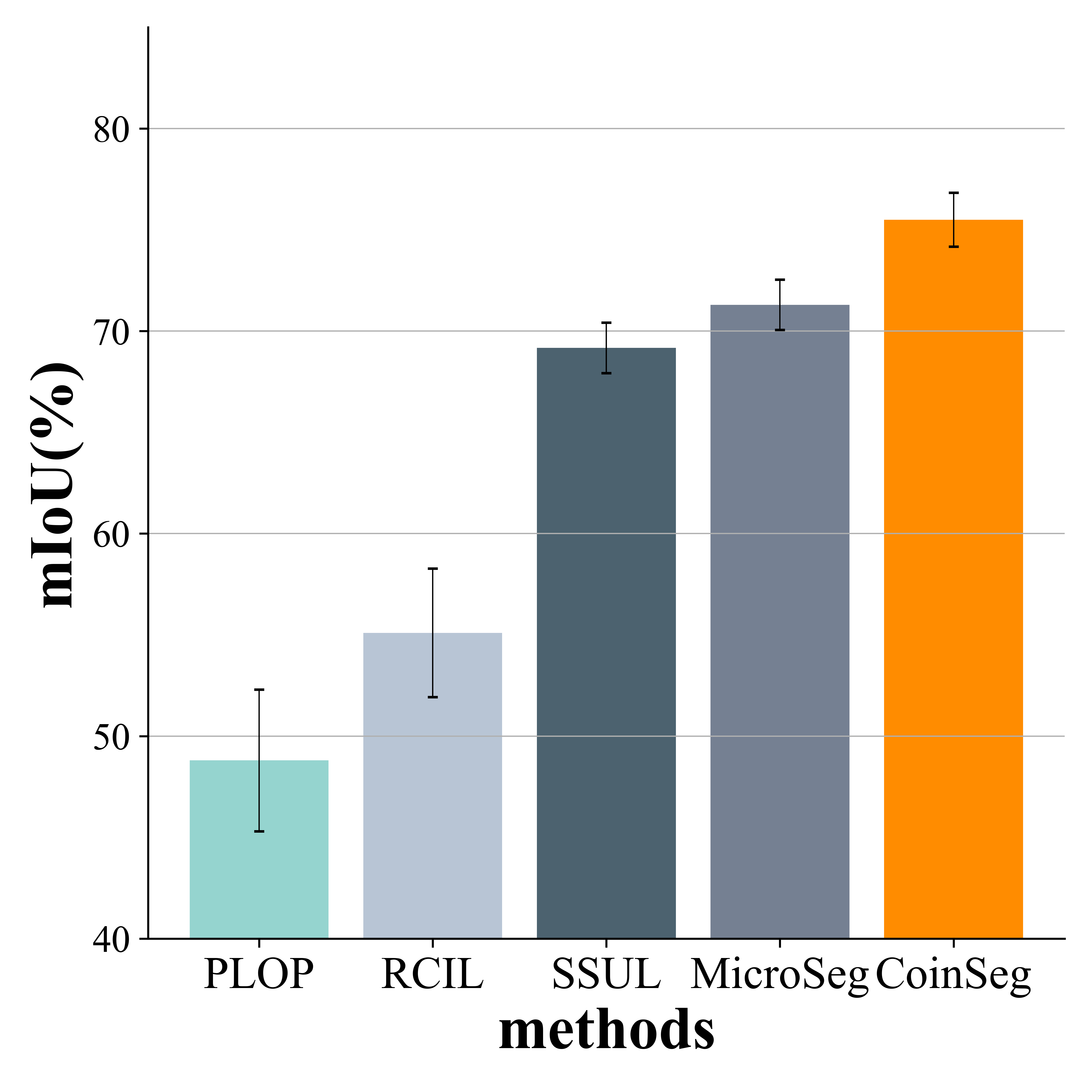} 
        \label{fig:memory_size}    
    }
\caption{ Performance comparisons for VOC. Illustration of the change of miou with learning step in (a) VOC 15-1, (b) VOC 2-2. Note we calculate mean IoU for all seen classes until current learning step. \eg, the mIoU of classes 0-10 for VOC 2-2 step 5. And (c) Average performance comparisons with 20 different incremental orders of VOC 15-1. 
    }
    \label{fig:compareVOC}
\end{figure*}

%% file: tables/ADE.tex
\begin{table*}[t]
  \centering
  \caption{
     Comparison with state-of-the-art methods on ADE20K. \dag: Re-implemented with Swin-B backbones; \textcolor{lightgray}{Joint} is the upperbound.
  }
  \label{tab:ade}
  \begin{adjustbox}{max width=\linewidth}
  \begin{tabular}{l|c|ccc|ccc|ccc|ccc}
    \toprule
    \multirow{2}{*}{Method} & \multirow{2}{*}{Backbone} & \multicolumn{3}{|c}{\textbf{ADE 100-5 (11 steps)}} & \multicolumn{3}{|c}{\textbf{ADE 100-10 (6 steps)}} & \multicolumn{3}{|c}{\textbf{ADE 100-50 (2 steps)}} & \multicolumn{3}{|c}{\textbf{ADE 50-50 (3 steps)}}  \\
    & & 0-100 & 101-150 & all & 0-100 & 101-150 & all & 0-100 & 101-150 & all & 0-50 & 51-150 & all    \\
    \midrule
    \textcolor{lightgray}{Joint} & \textcolor{lightgray}{Resnet101} & \textcolor{lightgray}{43.8} & \textcolor{lightgray}{28.9} & \textcolor{lightgray}{38.9} & \textcolor{lightgray}{43.8} & \textcolor{lightgray}{28.9} & \textcolor{lightgray}{38.9} & \textcolor{lightgray}{43.8} & \textcolor{lightgray}{28.9} & \textcolor{lightgray}{38.9} & \textcolor{lightgray}{50.7} & \textcolor{lightgray}{32.8} & \textcolor{lightgray}{38.9}  \\
    ILT~\cite{ILT_michieli2019incremental}& Resnet101 & 0.1 & 1.3 & 0.5 & 0.1 & 3.1 & 1.1 & 18.3 & 14.4 & 17.0 & 3.5 & 12.9 & 9.7 \\
    MiB~\cite{MiB}& Resnet101 & 36.0 & 5.7 & 26.0 & 38.2 & 11.1 & 29.2 & 40.5 & 17.2 & 32.8 & 45.6 &  21.0 & 29.3  \\
    PLOP~\cite{PLOP}& Resnet101 & 39.1 & 7.8 & 28.8 &  40.5 & 13.6 & 31.6 &  41.9 & 14.9 & 32.9 & 48.8 & 21.0 &  30.4  \\

    SSUL~\cite{SSUL}& Resnet101 &  39.9 &  17.4 &  32.5 & 40.2 &  18.8 &  33.1 & 41.3 & 18.0 &  33.6 & 48.4 & 20.2 & 29.6 \\
    RCIL~\cite{RCIL}& Resnet101 &  38.5 &  11.5 &  29.6 & 39.3 &  17.7 &  32.1 & \textbf{42.3} & 18.8 &  34.5 & 48.3 & 24.6 & 32.5 \\
    MicroSeg~\cite{MicroSeg} & Resnet101 & 40.4 & 20.5 & 33.8 & 41.5 & 21.6 & 34.9 & 40.2 & 18.8 & 33.1 & 48.6 & 24.8 & 32.9 \\

    \midrule
    \textcolor{lightgray}{Joint~\dag} & \textcolor{lightgray}{Swin-B} & \textcolor{lightgray}{43.5} & \textcolor{lightgray}{30.6} & \textcolor{lightgray}{39.2} & \textcolor{lightgray}{43.5} & \textcolor{lightgray}{30.6} & \textcolor{lightgray}{39.2} & \textcolor{lightgray}{43.5} & \textcolor{lightgray}{30.6} & \textcolor{lightgray}{39.2} & \textcolor{lightgray}{50.2} & \textcolor{lightgray}{33.7} & \textcolor{lightgray}{39.2}  \\
    SSUL~\dag~\cite{SSUL} & Swin-B & 41.3 & 16.0 & 32.9 & 40.7 & 19.0 & 33.5 & 41.9 & 20.1 & 34.6 & 49.5 & 21.3 & 30.7 \\

    MicroSeg~\dag~\cite{MicroSeg} & Swin-B & 41.2 & 21.0 & 34.5 & 41.0 & 22.6 & 34.8 & 41.1 & 24.1 & 35.4 & \textbf{49.8} & 23.9 & 32.5 \\

    \midrule 
    \methodname & Swin-B & \textbf{43.1} & \textbf{24.1} & \textbf{36.8} & \textbf{42.1} & \textbf{24.5} & \textbf{36.2} & {41.6} & \textbf{26.7} & \textbf{36.6} & 49.0 & \textbf{28.9} & \textbf{35.6}
    \\

    \bottomrule
  \end{tabular}
  \end{adjustbox}
  \vspace{-2mm}
\end{table*}

%% file: tables/ourmethod_M.tex
\begin{table*}[htp!]
  \centering
  \caption{
     Comparisons of CoinSeg using memory sampling strategy.  \dag: Re-implemented with Swin-B backbone.
  }
  \label{tab:abl_M}
  \begin{adjustbox}{max width=\linewidth}
     \begin{tabular}{l|c|ccc|ccc|ccc|ccc|ccc}
\toprule
\multirow{2}{*}{Method} & \multirow{2}{*}{Backbone} & \multicolumn{3}{c|}{\textbf{VOC 10-1 (11 steps)}} & \multicolumn{3}{c|}{\textbf{VOC 15-1 (6 steps)}} & \multicolumn{3}{c|}{\textbf{VOC 19-1 (2 steps)}} & \multicolumn{3}{c|}{\textbf{VOC 15-5 (2 steps)}} & \multicolumn{3}{c}{\textbf{VOC 2-2 (10 steps)}} \\
 &  & 0-10 & 11-20 & all & 0-15 & 16-20 & all & 0-19 & 20 & all & 0-15 & 16-20 & all & 0-2 & 3-20 & all \\ \midrule

SSUL-M~\cite{SSUL} & Resnet101 & 74.0 & 53.2 & 64.1 & 78.4 & 49.0 & 71.4 & 77.8 & 49.8 & 76.5 & 78.4 & 55.8 & 73.0 & 58.8 & 45.8 & 47.6 \\

MicroSeg-M~\cite{MicroSeg} & Resnet101 & 77.2 & 57.2 & 67.7 & 81.3 & 52.5 & 74.4 & 79.3 & 62.9 & 78.5 & 82.0 & 59.2 & 76.6 & 60.0 & 50.9 & 52.2 \\ \midrule

SSUL-M\dag~\cite{SSUL} & Swin-B & 75.3 & 54.1 & 65.2 & 78.8 & 49.7 & 71.9 & 78.5 & 50.0 & 77.1 & 79.3 & 55.1 & 73.5 & 61.1 & 47.5 & 49.4 \\

MicroSeg-M\dag~\cite{MicroSeg} & Swin-B & 78.9 & 59.2 & 70.1 & 82.0 & 47.3 & 73.7 & 81.0 & \textbf{62.4} & 80.0 & 82.9 & 60.1 & 77.5 & 62.7 & 51.4 & 53.0  \\ \midrule
\methodname~(Ours) & Swin-B & 80.0 & 63.4 & 72.5 & 82.7 & 52.5 & 75.5 & 81.5 & 44.8 & 79.8 & 82.1 & 63.2 & 77.6 & \textbf{70.1} & 63.3 & 64.3 \\ 
CoinSeg-M~(Ours) & Swin-B & \textbf{81.3} & \textbf{64.4} & \textbf{73.7} & \textbf{84.1} & \textbf{65.6} & \textbf{79.6} & \textbf{82.7} & 52.6 & \textbf{81.3} & \textbf{84.1} & \textbf{69.9} & \textbf{80.8} & 68.4 & \textbf{65.6} & \textbf{66.0} \\ 
\bottomrule
\end{tabular}
  
  \end{adjustbox}
  \vspace{-2mm}
\end{table*}

%% file: tables/detailed_ablation.tex
\begin{table}[t]
  \centering
  \caption{
     Ablation Studies for our proposed methods. \textit{Coin}: contrast inter- and intra-class representations, Fz: Freeze strategy, FLR: flexible initial learning rate. Numbers in the brackets (): gains w.r.t. the preceding row.
  }
  \label{tab:detailed_abl}
  \begin{adjustbox}{max width=\linewidth}
     \begin{tabular}{c|c|cc|cc|ccc}
     \toprule
 \multirow{2}{*}{{Backbone}}  & \multicolumn{3}{c|}{parameter strategy} & \multicolumn{2}{c|}{\textit{Coin}} & \multicolumn{3}{c}{\textbf{VOC 15-1 (6 steps)}} \\
 & LR & $\mathcal{L}^{F}_{kd}$ & $\mathcal{L}^{z}_{kd}$ & $\mathcal{L}_{int}$ & $\mathcal{L}_{itr}$ & 0-15 & 16-20 & all \\
 \midrule
ResNet101 & Fz & \xmark & \xmark & \xmark & \xmark & 74.9 & 26.4 & 63.3 \\
Swin-B & Fz & \xmark & \xmark & \xmark & \xmark & 79.5 & 42.4 & 70.5 \\
\midrule
Swin-B & FLR & \xmark & \xmark & \xmark & \xmark & 73.6 & 44.0 & 66.7 \\
Swin-B & FLR & \cmark & \xmark & \xmark & \xmark & 78.9 & 42.7 & 70.3~(+3.6) \\
Swin-B & FLR & \cmark & \cmark & \xmark & \xmark & 80.4 & 43.7 & 71.6~(+1.3) \\
Swin-B & FLR & \cmark & \cmark & \cmark & \xmark & 80.8 & 45.9 & 72.4~(+0.8) \\
Swin-B & FLR & \cmark & \cmark & \cmark & \cmark & 82.7 & 52.5 & \textbf{75.5}~(+3.1)
\\
\bottomrule
\end{tabular}

  \end{adjustbox}
\end{table}

%% file: tables/ablation_proposal_N.tex
\begin{table}[t]

  \centering
  \caption{
     Ablations to \# of proposals ($N$) in $\mathcal{L}_{itr}$~(Left), and pseudo-labeling  in $\mathcal{L}_{int}$~(Right). 
     GT: ground truth, PL: pseudo label.
  }
  \label{tab:proposal_N_abl}
     \begin{tabular}{cc|ccc}
      \toprule
      \multirow{2}{*}{{Method}} &\multirow{2}{*}{$N$} & \multicolumn{3}{c}{\textbf{VOC 15-1 (6 steps)}}          \\
                          &                                       & 0-15          & 16-20         & all \\ 
                          \midrule
       \multirow{3}{*}{CoinSeg}                
        & 50 & 81.4 & 51.1 & 74.2         \\
                                                            
        & 100 & 82.7 & 52.5 & 75.5  \\
        & 200 & 83.1  & 53.8   & \textbf{76.1}  \\
      \bottomrule
    \end{tabular}
  
  \vspace{-2mm}
\end{table}

%% file: figures/visualization.tex
\begin{figure*}[t]
\centering 
{\includegraphics[width=1.00\linewidth]{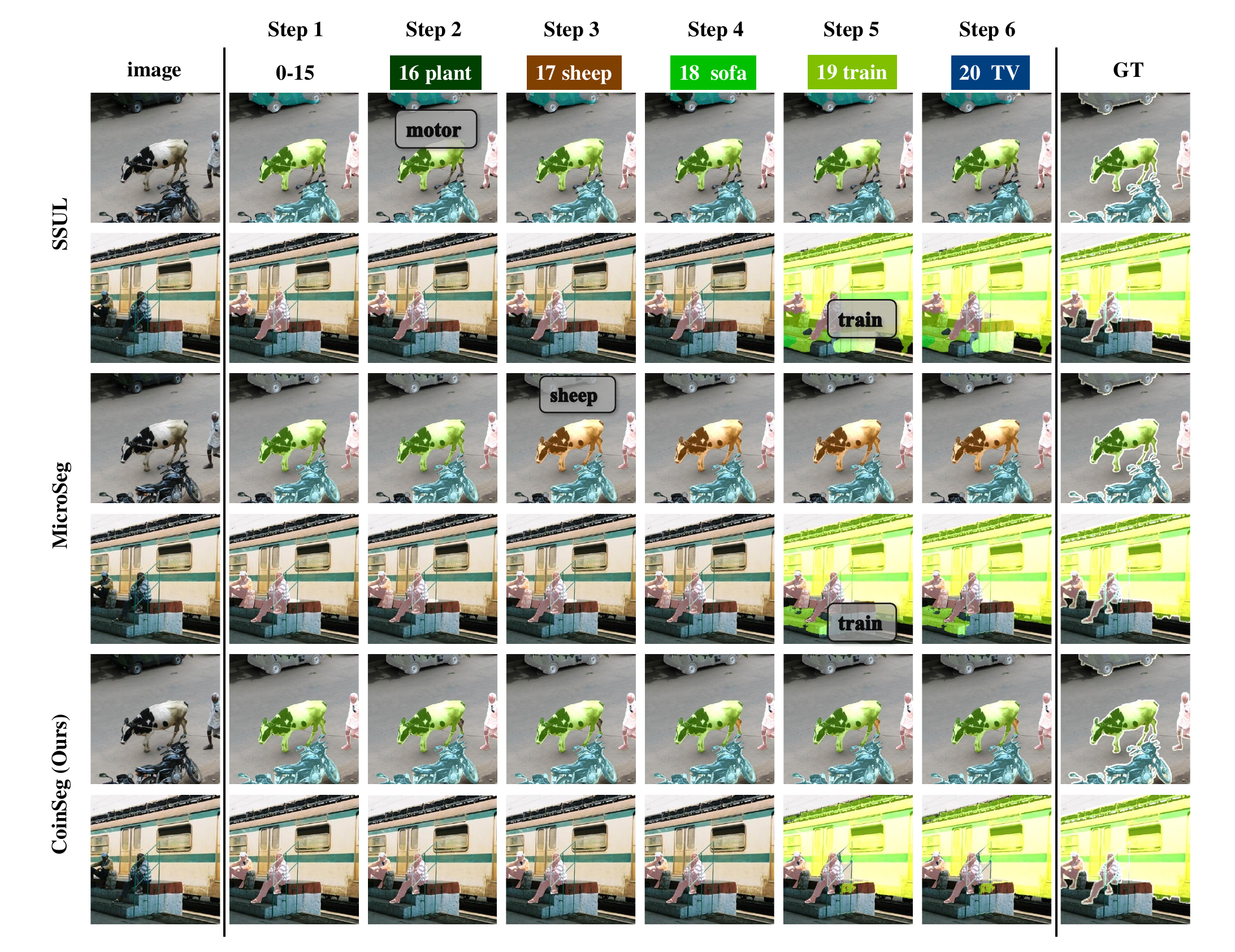}}
\vspace{-5mm}
\caption{{Qualitative analysis for VOC, comparing with prior CISS methods. Text points out the \textbf{uncorrectly predicted} areas.}}
\label{figure:vis}
 \end{figure*}

%% file: tables/expandion.tex
\begin{table}[t]
  \centering
  \caption{
     The performance of Flexible tuning for prior freeze-strategy-based method MicroSeg, `FT' denotes flexible tuning .
  }
  \label{tab:abl_FL}
  \begin{adjustbox}{max width=0.95\linewidth}
     \begin{tabular}{c|c|c|ccc}
      \toprule

        \multirow{2}{*}{{Method}}  & \multirow{2}{*}{{Backbone}}        & parameter & \multicolumn{3}{c}{\textbf{VOC 15-1~(6 steps)}}                                 \\
         & & strategy            & 0-15                                & 16-20         & all           \\ \midrule
      \multirow{4}{*}{MicroSeg} & Resnet101  & Freeze     & 80.5    & 40.8       & 71.0          \\
      & Resnet101 & FT   & 80.9       & 41.5 & \textbf{72.3}~(+1.3) \\
      
         & Swin-B & Freeze            & 80.1      & 36.8    & 69.8          \\
       & Swin-B & FT   & 79.8      & 40.2 & \textbf{70.4}~(+0.6) \\
      \bottomrule
    \end{tabular}
  
  \end{adjustbox}
  \vspace{-2mm}
\end{table}

%% file: sections/5_conclusion.tex
\section{Conclusion}

In this work, we studied class incremental semantic segmentation and proposed an effective method CoinSeg. Inspired by the Gaussian mixture model, we proposed \textit{Coin} to better characterize samples with explicitly constrain inter- and intra- class diversity. Furthermore, we proposed a flexible tuning strategy, to keep the stability of the model and alleviate forgetting by the flexible initial learning rate and regularization constraints. 
Extensive experimental evaluations show the effectiveness of our method. CoinSeg outperforms prior state-of-the-art CISS methods, especially on more challenging long-term incremental scenarios.

\paragraph{Acknowledgment.} 
This work was supported in part by the National Key R\&D Program of China (No.2021ZD0112100), the National Natural Science Foundation of China~(No. 61972036), the Fundamental Research Funds for the Central Universities (No. K22RC00010).